\documentclass{article}
\usepackage{ijcai18}
\usepackage{enumerate} 
\usepackage{times}
\usepackage[usenames,dvipsnames,svgnames,table]{xcolor}
\usepackage{soul}
\usepackage[utf8]{inputenc}
\usepackage[small]{caption}
   \usepackage{mdframed}
  \usepackage{graphicx}
  \usepackage{verbatim}
  \usepackage{fancybox}
  \usepackage{amssymb}
  \usepackage{url}
  \usepackage{amsmath}
  \usepackage{proof}  
  \usepackage{bussproofs}
  \usepackage{verbatim}
  \usepackage{algorithm2e}
  \usepackage{hyperref}
    \hypersetup{
      colorlinks,
      citecolor=blue,
      filecolor=blue,
      linkcolor=blue,
      urlcolor=blue
                 }
  \usepackage{paralist}
  \usepackage[]{algorithm2e}
  \usepackage{graphicx}
  \usepackage{subcaption}

\usepackage{booktabs}
\usepackage{color}
\usepackage{amsfonts}
\usepackage{amsmath}
\usepackage{proof}
\usepackage[amsmath,thmmarks]{ntheorem}
\usepackage{relsize}
 
\usepackage{mathtools} 

\usepackage{colortbl}

 \newcommand{\citeaffixed}[1]{\cite{#1}}

\newcommand{\IGNORE}[1]{}

\newcommand{\citeasnoun}[1]{%
  \citeauthor{#1}~[\citeyear{#1}]}

\newcommand{\lsort}[1]{%
  \ensuremath{\mbox{\textsf{#1}}}}
\newcommand{\defsort}[2]{%
  \newcommand{#1}{\lsort{#2}}}
\defsort{\Action}{Action}
\defsort{\Time}{Time}
\defsort{\Self}{Self}

\defsort{\Agent}{Agent}
\defsort{\Entrant}{Entrant}
\defsort{\ActionType}{ActionType}
\defsort{\Moment}{Moment}
\defsort{\Boolean}{Formula}
\defsort{\PayOut}{PayOut}
\defsort{\Fluent}{Fluent}
\defsort{\Event}{Event}
\defsort{\Object}{Object}
\defsort{\RealTerm}{RealTerm}

\defsort{\Numeric}{Numeric}
\defsort{\Number}{Number}
\defsort{\Trolley}{Trolley}
\defsort{\Track}{Track}
\defsort{\Moveable}{Moveable}

\newcommand{\lsymbol}[1]{%
  \ensuremath{\mathit{#1}}}
\newcommand{\defsymbol}[2]{%
  \newcommand{#1}{\lsymbol{#2}}}
\defsymbol{\actionType}{action}
\defsymbol{\initially}{initially}
\defsymbol{\holds}{holds}
\defsymbol{\happens}{happens}
\defsymbol{\clipped}{clipped}
\defsymbol{\initiates}{initiates}
\defsymbol{\terminates}{terminates}
\defsymbol{\prior}{prior}
\defsymbol{\interval}{interval}

\defsymbol{\does}{does}
\defsymbol{\plans}{plans}
\defsymbol{\act}{act}
\defsymbol{\react}{react}
\defsymbol{\payTot}{pay_{tot}}
\defsymbol{\fight}{fight}
\defsymbol{\coop}{coop}
\defsymbol{\enter}{enter}
\defsymbol{\stayout}{stayout}
\defsymbol{\learns}{learns}
\defsymbol{\payoff}{payoff}
\defsymbol{\position}{position}
\defsymbol{\dead}{dead}
\defsymbol{\damaged}{damaged}

\defsymbol{\onrails}{onrails}
\defsymbol{\switch}{switch}
\defsymbol{\drop}{drop}
\defsymbol{\plan}{plan}

\newcommand{\lconstant}[1]{%
  \ensuremath{\mbox{\textsf{#1}}}}
\newcommand{\defconstant}[2]{%
  \newcommand{#1}{\lconstant{#2}}}
\defconstant{\Enter}{Enter}
\defconstant{\StayOut}{StayOut}
\defconstant{\Fight}{Fight}
\defconstant{\Acquiesce}{Acquiesce}
\defconstant{\cs}{cs }

\newcommand{\lmodality}[1]{%
  \ensuremath{\mathbf{#1}}}
\newcommand{\defmodality}[2]{%
  \newcommand{#1}{\lmodality{#2}}}
\defmodality{\common}{C}
\defmodality{\knows}{K}
\defmodality{\believes}{B}
\defmodality{\perceives}{P}
\defmodality{\mental}{M}

\defmodality{\desires}{D}
\defmodality{\intends}{I}
\defmodality{\says}{S}
\defmodality{\ought}{O}

\newcommand{\sep}{\ \lvert \ }

\usepackage{booktabs}
\usepackage{mdframed}
\usepackage{paralist}

\usepackage{color}

\newcommand{\DDE}{\ensuremath{\mathcal{{DDE}}}}
\newcommand{\DCEC}{\ensuremath{{\mathcal{{DCEC}}}}}

\newcommand{\type}[1]{\textsf{#1}}

\newcommand{\Believes}{\ensuremath{\mathbf{B}}}

\newcommand{\Knows}{\ensuremath{\mathbf{K}}}

\newcommand{\can}{\ensuremath{\can}}

\providecommand{\shortcite}[1]{\cite{#1}}

\newcommand{\level}[2]{\ensuremath{\mathsf{Level}}(#1)}

\title{Tentacular Artificial Intelligence, and the Architecture Thereof, Introduced}

\author{}

\author{
\small{Selmer Bringsjord}$^1$, 
Naveen Sundar G$^1$, 
Atriya Sen$^1$, 
Matthew Peveler$^1$, 
Biplav Srivastava$^2$
Kartik Talamadupula$^2$
\\ 
$^1$ \small{Rensselaer Polytechnic Institute (RPI); RAIR Lab}\\
$^2$ \small{IBM Research}\\
\small{\texttt{Selmer.Bringsjord@gmail.com}},
\small{\texttt{naveensundarg@gmail.com}},
\small{\texttt{atriya@atriyasen.com}},\\
\small{\texttt{matt.peveler@gmail.com}},
\small{\texttt{biplavs@us.ibm.com}},
\small{\texttt{krtalamad@us.ibm.com}}
}

\begin{document}

\maketitle

\begin{abstract}
\noindent
  We briefly introduce herein a new form of distributed, multi-agent
  artificial intelligence, which we refer to as ``tentacular.''
  Tentacular AI is distinguished by six attributes, which among other
  things entail a capacity for reasoning and planning based in highly
  expressive calculi (logics), and which enlists subsidiary agents
  across distances circumscribed only by the reach of one or more
  given networks.
\end{abstract}

\section{Introduction} 
We briefly introduce herein a new form of distributed, multi-agent
artificial intelligence.  An AI artifact $s$ is currently understood
as an agent with a predetermined set of goals, a set of fixed inputs
and outputs, and obligations and permissions.  The agent does not have
any leeway in accomplishing its goals or adhering to its obligations,
prohibitions, or other legal/ethical principles that bind it.  Do we
need agents that go beyond these limitations?  A humble example
follows: During your daily commute to work, an agent $a_c$ in your car
observes that there is more traffic than usual headed toward the local
store.  It then consults a weather service and finds that a major
storm is headed toward your town.  $a_c$ conveys this information to
$a_h$, an agent in your home.  $a_h$ then communicates with an agent
$a_p$ on your phone and finds out that you do not know about the storm
coming your way, as you have not made any preparations for it; and as
further evidence of your ignorance, you have not read any
notifications about the storm.  $a_h$ then infers from your calendar
that you may not have enough time to get supplies after you read your
notifications later in the day.  $a_h$ commands $a_c$ to recommend to
you a list of supplies to shop for on your way home, including at
least $n$ items in certain categories (e.g.\ 3 gallons of bottle
water).

AI of today, as defined by any orthodox, comprehensive overview of it
(e.g.\ \cite{aima.third.ed}), consists in the design, creation,
implementation, and analysis of \textbf{artificial
  agents}.\footnote{This is the exact phrase used by
  \citeasnoun{aima.third.ed}.  Other comprehensive overviews match the
  Russell-Norvig orientation; e.g.\ \cite{luger_ai_book_6thed}.} Each
such agent $a$ takes in information about its particular environment
$E$ (i.e.\ takes in \textbf{percepts} of $E$), engages in some
computation, and then, on the strength of that computation, performs
an action/actions in that environment.  (Of course, for an agent that
persists, this cycle iterates through time.)  On this definition, a
computer program that implements, say, the factorial function $n!$
qualifies as an artificial agent (let's dub it
`$a_{\scriptstyle{\textsc{fac}}}$'), one operating in the environment
$\mathbf{E}_{\scriptstyle{\mathcal{N}}}$ of basic arithmetic; and the
human who has conceived and written this program has built an
artificial agent.  While plenty of the artificial agents touted today
are rather more impressive than $a_{\scriptstyle{\textsc{fac}}}$, our
aim is to bring to the world, within a decade, a revolutionary kind of
AI that yields artificial agents with a radically higher level of
intelligence (including intelligence high enough to qualify the
agents as \textit{cognitively conscious}) and power.  This envisioned
AI we call \textbf{Tentacular AI}, or just `TAI' for short (rhymes
with `pie').  Before presenting architectural-level information about
TAI, we give an example that's a bit more robust than our
first-paragraph one.

Let's suppose that an AI agent $a_{\scriptstyle{\textsc{home}}}$
overseeing a home is charged with the single, unassuming task of
moving a cup on the home's kitchen table onto a saucer that is also on
that table.  How shall the agent make this goal happen?  If the AI can
delegate to a robot in the house capable of manipulating standard
tabletop objects in a narrow tabletop environment
$\mathbf{E}_{\scriptstyle{\textsc{table}}}$, and that robot is at the
table or can get there in a reasonable amount of time, then of course
$a_{\scriptstyle{\textsc{home}}}$ can direct the robot to pick up
the cup and put it on the saucer.  This is nothing to write home
about, since AI of today has given us agent-robot combos that, in labs
(our own, e.g.)~and soon enough in homes across the technologized
world, can do this kind of thing routinely and reliably.  In fact,
this kind of capability to find plans and move tabletop objects around
in order to obtain goals in tabletop environments\footnote{Such
  environments are variants of those traditionally termed
  `blocks-worlds.'} has been a solved problem from the research point
of view for decades \citeaffixed{logical.foundations.ai}, e.g.
Not only that, but there are longstanding theorems telling us that the
intrinsic difficulty of finding plans to move various standard
tabletop objects in arbitrary starting configurations in tabletop
environments is algorithmically solvable and generally
tractable.\footnote{E.g., see \cite{blocks-world_complexity}.}

However, AI of today is, if you will, living a bit of a lie.  Why?
Because in real life, the agent $a_{\scriptstyle{\textsc{home}}}$
would \emph{not} be operating in only the tabletop environment
$\mathbf{E}_{\scriptstyle{\textsc{table}}}$; rather the idea is that
this agent should be able to understand and manage the overall
environment $\mathbf{E}_{\scriptstyle{\textsc{home}}}$ of the home,
which surely comprises much more than the stuff standardly on one
kitchen table!  Homes can have parents, kids, dogs, visitors, $\ldots$
\textit{ad indefinitum}.

For example, suppose that $a_{\scriptstyle{\textsc{home}}}$ finds
that the tabletop robot is broken, having been mangled by the home's
frisky beagle.  \emph{Then} how does
$a_{\scriptstyle{\textsc{home}}}$ solve the problem of getting
the saucer moved?  Artificial agents of today capable of the kind of
planning that worked before this complication are now hamstrung.  But
not so a TAI agent.  One reason is that TAI agents are capable of
human-level communication.  In certain circumstances within
$a_{\scriptstyle{\textsc{home}}}$ the most efficient way for the
agent $a_{\scriptstyle{\textsc{home}}}$ to accomplish the task
may be to simply say politely via I/IoT (Internet or Interent of Things) through a speaker or a
smartphone or a pair of smart glasses to a human in the home (of whose
mind the TAI agent has a model) sitting at the table in question:
``Would you be so kind as to place that cup on top of the saucer?''
Of course, $a_{\scriptstyle{\textsc{home}}}$ may not be so
fortunate as to have the services of a human available: maybe no human
is at home, yet the task must be completed.  In this case, a TAI agent
can still get things done, in creative fashion.  E.g.,~suppose that in
the home a family member received beforehand a small blimp that can
fly around inside the home and pick things up.\footnote{Such a blimp
  is a simple adaptation of what is readily available as a relatively
  inexpensive toy.}  The TAI agent might then activate and use this
blimp through I/IoT to put the cup atop the saucer.  But what, more
precisely, is a TAI agent?
We say that a TAI agent must be:

\begin{small}
    \begin{enumerate}

      \item[$\mathbf{D_1}$] \textit{Capable of problem-solving}.  Whereas, as we've
        noted, standard AI counts simple mappings from percepts to
        actions as \textit{bona fide} AI, TAI agents must be capable
        of problem-solving.  This may seem like an insignificant first
        attribute of TAI, but a consequence that stems from this
        attribute should be noted: Since problem-solving entails
        capability across the main sub-divisions of AI, TAI agents
        have multi-faceted power.  Problem-solving requires capability
        in these sub-areas of AI: planning, reasoning, learning,
        communicating, creativity (at least relatively simple forms
        thereof), and --- for making physical changes in physical
        environments --- cognitive robotics.\footnote{Cognitive
          robotics is defined in \cite{Levesque07cognitiverobotics} as
          a type of robotics in which all substantive actions
          performed by the robots are a function of the cognitive
          states (e.g.\ beliefs \& intentions) of these robots.}
        Hence, all TAI agents can plan, reason, learn, communicate;
        and they are creative and capable of carrying out physical
        actions.

      \item[$\mathbf{D_2}$]  \textit{Capable of solving at least important instances of
        problems that are at and/or \textbf{above} Turing-unsolvable
        problems}.  AI of today, when capable of solving problems,
        invariably achieves this success on problems that are merely
        algorithmically solvable and tractable (e.g., checkers, chess,
        Go).

      \item[$\mathbf{D_3}$]  \textit{Able to supply justification, explanation, and
        certification of supplied solutions, how they were arrived at,
        and that these solutions are safe/ethical}.  We thus say that
        the problem-solving of a TAI agent is \textbf{rationalist}.
        This label reflects the requirement that any proposed solution
        to the problem discovered by a TAI agent must be accompanied
        by a justification that defends and explains that the proposed
        solution \emph{is} a solution, and, when appropriate, also
        that the solution (and indeed perhaps the process used to
        obtain the solution) has certain desirable properties.
        Minimally, the justification must include an argument or proof
        for the relevant conclusions.  In addition, the justification
        must be verified, formally; we thus say that
        \textbf{certification} is provided by a TAI agent.

      \item[$\mathbf{D_4}$] \textit{Capable of ``theory-of-mind''
        level reasoning, planning, and communication}.  Discussion of
        this attribute is omitted to save space; see
        e.g.\ \cite{ka_sb_scc_seqcalc} for our lab's first foray into
        automated reasoning at this level.  (The truth is, it's more
        accurate to say the fourth requirement is that a TAI agent
        must have \textit{cognitive consciousness}, as this phenomenon
        is explained and axiomatized in
        \cite{axiomatizing_consciousness1}.)

      \item[$\mathbf{D_5}$] \textit{Capable of creativity, minimally
        to the level of so-called \textbf{m-creativity}}.  Creativity
        in artificial agents, and the engineering thereof, has been
        discussed in a number of places by Bringsjord
        \citeaffixed{brutus}{e.g.}, but recently
        \citeasnoun{creative_cars} have called for a form of
        creativity in artificial agents using I/IoT.

      \item[$\mathbf{D_6}$] \textit{Has ``tentacular'' power wielded
          throughout I/IoT, Edge Computing, and cyberspace}.  This is
        the most important attribute possessed by TAI agents, and is
        reflected in the `T' in `TAI.' To say that such agents have
        tentacular problem-solving power is to say that they can
        perceive and act through the I/IoT (or equivalent networks)
        and cyberspace, across the globe.  TAI agents thus operate in
        a planet-sized, heterogeneous environment that spans the
        narrower, fixed environments used to define conventional,
        present-day AI, such as is found in \cite{aima.third.ed}.

    \end{enumerate}
\end{small}

\section{Related Work}
Given the limited scope of the present paper, we only make some brief
comments about related work, which can be partitioned for convenience
into that which is can be plausibly regarded as on the road toward the
level of expressivity and associated automated reasoning that TAI
requires, and prior work that provides a stark and illuminating
contrast with TAI.

First, as to work we see as reaching toward TAI, we note that recently
\citeauthor{Miller2018} [\citeyear{Miller2018}] present a planning
framework that they call \emph{social planning}, in which the agent
under consideration can plan and act in a manner that takes account of
the beliefs of other agents.  The goal for an agent in social planning
can either be a particular state of the external world, or a set of
beliefs of other agents (or a mix of both).  The system is built upon
a simplified version of a propositional modal logic (unlike our
system, presented below, which is more expressive and can accommodate
more complex goals, e.g.\ goals over unbounded domains or goals that
involve numerical quantification; such statements require going beyond
propositional modal logic).  In addition, certainly the NARS system
from Wang [\citeyear{nars_rigid_flexibility}] has elements that one
can rationally view as congenial to TAI.  For instance, NARS is
multi-layered and reasoning-centric.  On the other hand, the `N' in
`NARS' is for `Non-axiomatic,' and TAI, and indeed the entire approach
to logicist AI pursued by at least Bringsjord and Govindarajulu, seeks
whenever possible to leverage automated reasoning over powerful axiom
systems, such as Peano Arithmetic.\footnote{The layering of TAI is in
  fact anticipated by the increasingly powerful axiom-centric
  cognition described in
  \cite{general_intelligence_implies_creativity}, which takes Peano
  Arithmetic as central.} In addition, TAI is deeply and irreducibly
intensional, while NARS appears to be purely extensional.  Clever
management of computational resources in TAI is clearly going to be
key, and we see the work of Thorisson and colleagues (e.g.\
\cite{attention_mechanisms_thorisson}) to be quite relevant to TAI and
the challenges the implementation of it will encounter.  For a final
example of work that is generally aligned with TAI, we bring to the
reader's attention a recent comprehensive treatment of proof-based
work in computer science: \cite{fpmics}.  As TAI is steadfastly
proof-based AI, this tome provides very nice coverage of the kind of
work required to implement TAI.

Secondly, for illuminating contrast, we note first that some have
considered the concept of corporate intelligence composed of multiple
agents, including machines, where inspiration comes from biology.  A
case in point is the fascinating modeling in
\cite{intelligence_campus_chella}.\footnote{Though out of reach for
  now, given that our chief objective is but an informative
  introduction to TAI, the relationship between our conception of
  cognitive consciousness, which is central to TAI agents (Attribute
  \#4 above), and consciousness as conceived by Chella, is a fertile
  topic for future investigation.  A multi-faceted discussion of
  artificial consciousness is by the way to be had in
  \cite{artificial_consciousness_chella}.  For a first-draft
  axiomatization of the brand of consciousness central to TAI agents,
  see \cite{axiomatizing_consciousness1}.}  In our case, TAI is a
thoroughly formal conception independent of terrestrial biology, one
that is intended to include types of agents of greater intelligence
than those currently on Earth.  Another illuminating contrast comes
via considering established languages for planning that are purely
extensional in nature (e.g.\ PDDL, which in its early form is given in
\cite{pddl1}), as therefore quite different than planning of the type
that is required for TAI, which must be intensional in character, and
is (since cognitive calculi are intensional computational logics).
MA-PDDL is an extension of PDDL for handling domains with multiple
agents with varying actions and goals \cite{kovacsmapddl2012}, and as
such would seem to be relevant to TAI.  But unlike social planning
discussed above, MA-PDDL does not aim to change beliefs (nor for that
matter other epistemic attitudes) of other agents.  While MA-PDDL
could be used to do so, representing beliefs and other cognitive
states in PDDL's extensional language can lead to undesirable
consequences, as demonstrated in
\cite{selmer_naveen_metaphil_web_intelligence}.  Extensions of the
original PDDL (PDDL1), for example PDDL3 \cite{pddl3}, are still
extensional in nature.

This concludes the related-work section.  Note that below we describe
and define TAI from the point of view of AI planning.


\section{Quick Overview}
\begin{figure}[h!]
 \centering
 {
  \includegraphics[width=\linewidth]{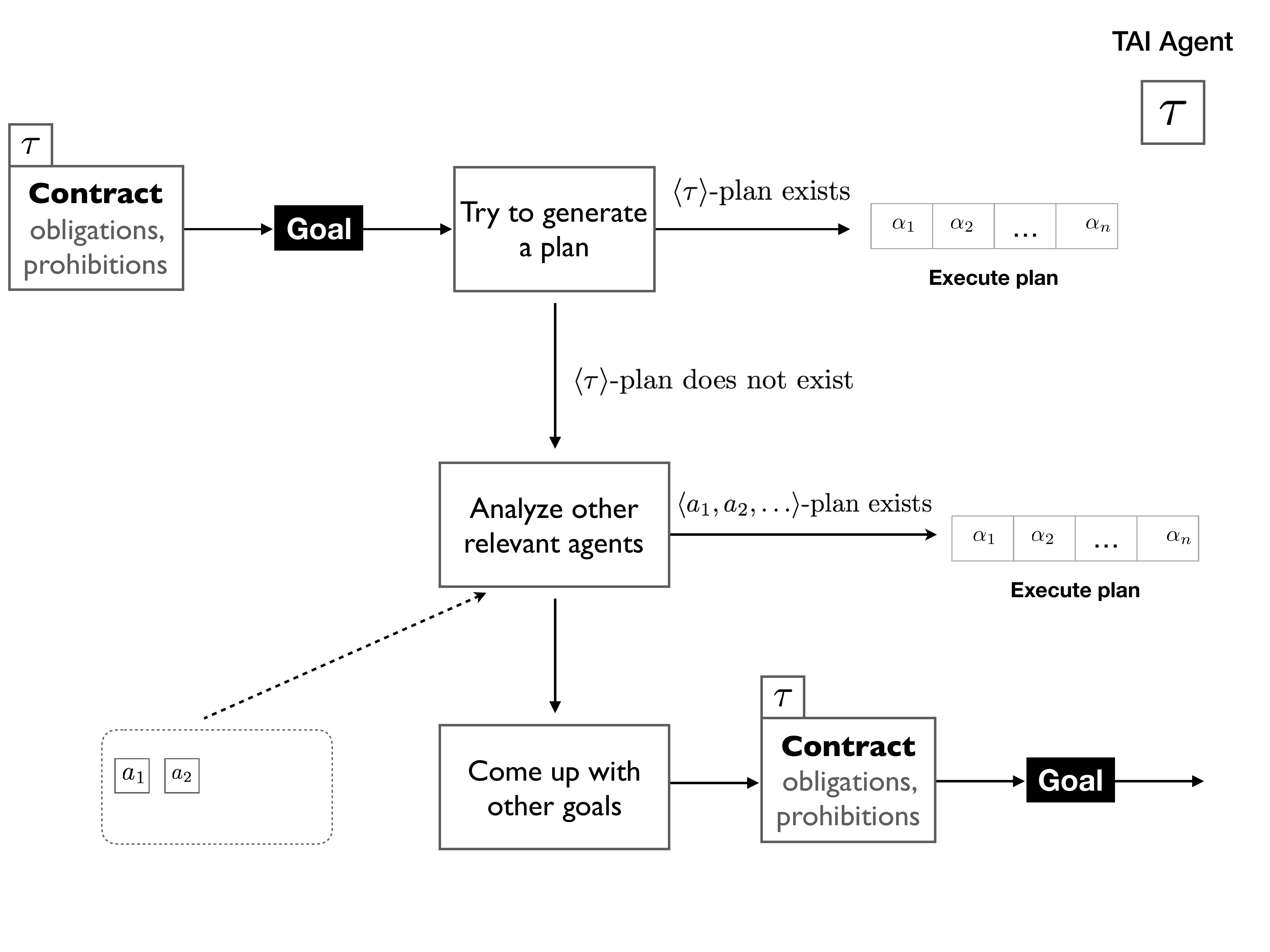}}
\caption{TAI Informal Overview: \textit{We have an architecture for
    how a TAI agent $\tau$ might operate. $\tau$ continuously comes up
    with goals based on its contract. If a goal is not achievable
    using $\tau$'s own resources, $\tau$ has to employ other agents in
    achieving this goal. To successfully do so $\tau$ would need to
    have one or more of $\mathbf{D_1} - \mathbf{D_6}$ attributes.  }}
 \label{fig: TAI_Flowchart}
\end{figure}

We give a quick and informal overview of TAI. We have a set of
agents $a_1, \ldots, a_n$. Each agent has an associated (implicit or
explicit) contract that it should adhere to. Consider one particular
agent $\tau$. During the course of this agent's lifetime, the agent
comes up with goals to achieve so that its contract is not
violated. Some of these goals might require an agent to exercise some
or all of the six attributes $\mathbf{D_1} - \mathbf{D_6}$. We
formalize this using planning as shown in Figure ~\ref{fig:
  TAI_Flowchart}. As shown in the figure, if some goal is not
achievable on its own, $\tau$ can seek to recruit other agents by
leveraging their resources, beliefs, obligations etc.


\section{The Formal System}

\begin{figure}[h!]
 \centering
 {
  \includegraphics[width=\linewidth]{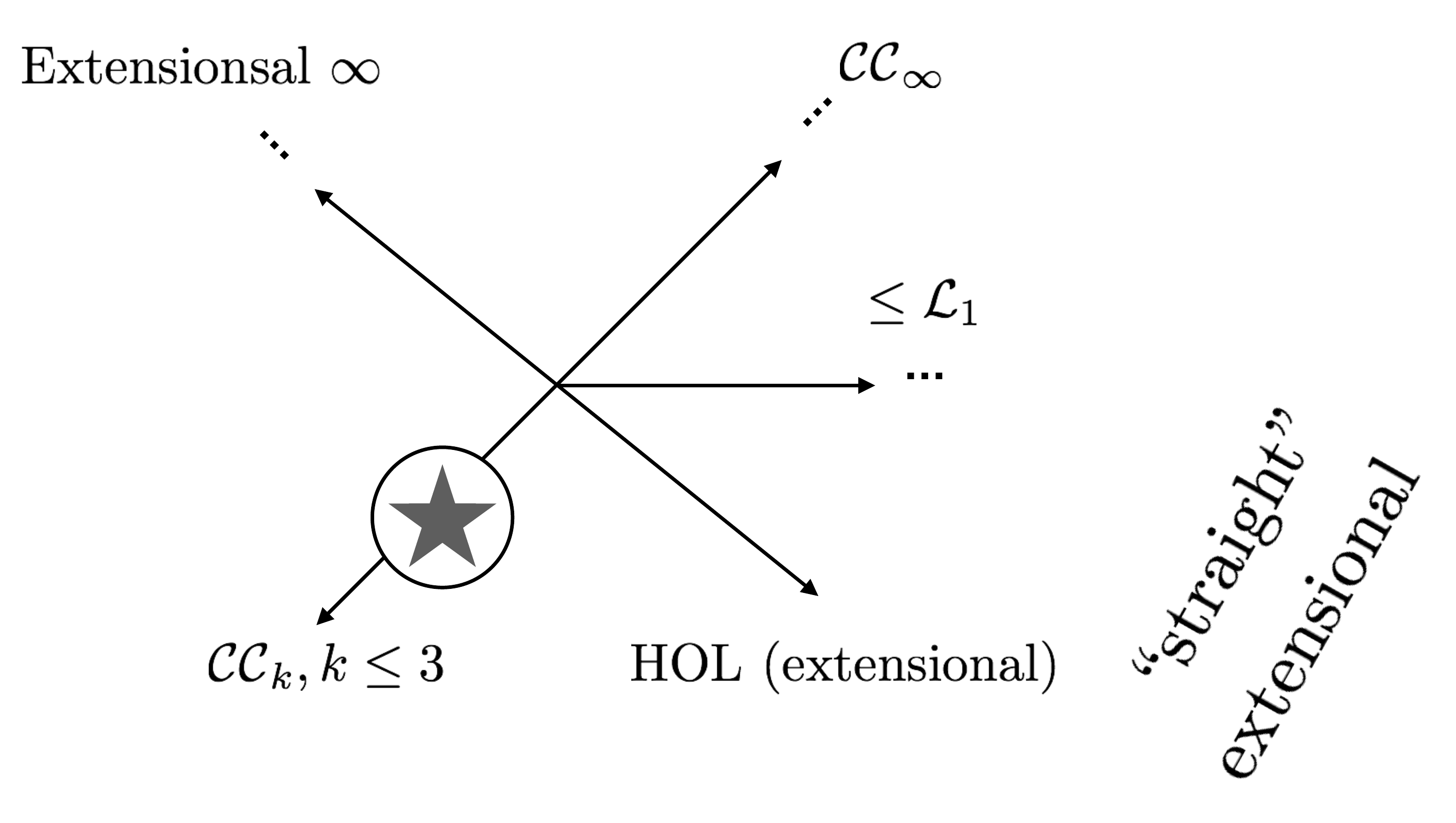}}
 \caption{Space of Logical Calculi.  \textit{There are five dimensions
     that cover the entire, vast space of logical calculi.  The due
     West dimension holds those calculi powering the Semantic Web
     (which are generally short of first-order logic =
     $\mathcal{L}_1$), and include so-called \textbf{description
       logics}.  Both NW and NE include logical systems with wffs that
     are allowed to be infinitely long, and are needless to say hard
     to compute with and over.  SE is higher-order logic, which has
     a robust automated theorem-proving community gathered around it.
     It's the SW direction that holds the cognitive calculi described
     in the present paper, and associated with TAI; and the star refers
     to those specific cognitive calculi called out in these pages by
     us.}}
 \label{fig:star_fig}
\end{figure}

To make the above notions more concrete, we use a version of a
computational logic.  The logic we use is \textbf{deontic cognitive
  event calculus} (\DCEC).  This calculus is a first-order modal
logic. Figure ~\ref{fig:star_fig} shows the region where \DCEC\ is
located in the overall space of logical calculi. \DCEC\ belongs to the
\textbf{cognitive calculi} family of logical calculi (denoted by a
star in Figure~\ref{fig:star_fig} and expanded in
Figure~\ref{fig:cc_family}). \DCEC\ has a well-defined syntax and
inference system; see Appendix A of \cite{nsg_sb_dde_2017} for a full
description. The inference system is based on natural deduction
\cite{gentzen_investigations_into_logical_deduction}, and includes all
the introduction and elimination rules for first-order logic, as well
as inference schemata for the modal operators and related structures

This system has been used previously in
\cite{nsg_sb_dde_2017,dde_self_sacrifice_2017} to automate versions of
the doctrine of double effect \DDE, an ethical principle with
deontological and consequentialist components.  While describing the
calculus is beyond the scope of this paper, we give a quick overview
of the system below.  Dialects of \DCEC\ have also been used to
formalize and automate highly intensional (i.e. cognitive) reasoning
processes, such as the false-belief task
\cite{ArkoudasAndBringsjord2008Pricai} and \textit{akrasia}
(succumbing to temptation to violate moral principles)
\cite{akratic_robots_ieee_n}. {Arkoudas and Bringsjord
  \shortcite{ArkoudasAndBringsjord2008Pricai} introduced the general
  family of \textbf{cognitive event calculi} to which \DCEC\ belongs,
  by way of their formalization of the false-belief task.} More
precisely, \DCEC\ is a sorted (i.e.\ typed) quantified modal logic
(also known as sorted first-order modal logic) that includes the event
calculus, a first-order calculus used for commonsense reasoning.

\begin{figure}[h!]
 \centering
 {
  \includegraphics[width=0.75\linewidth]{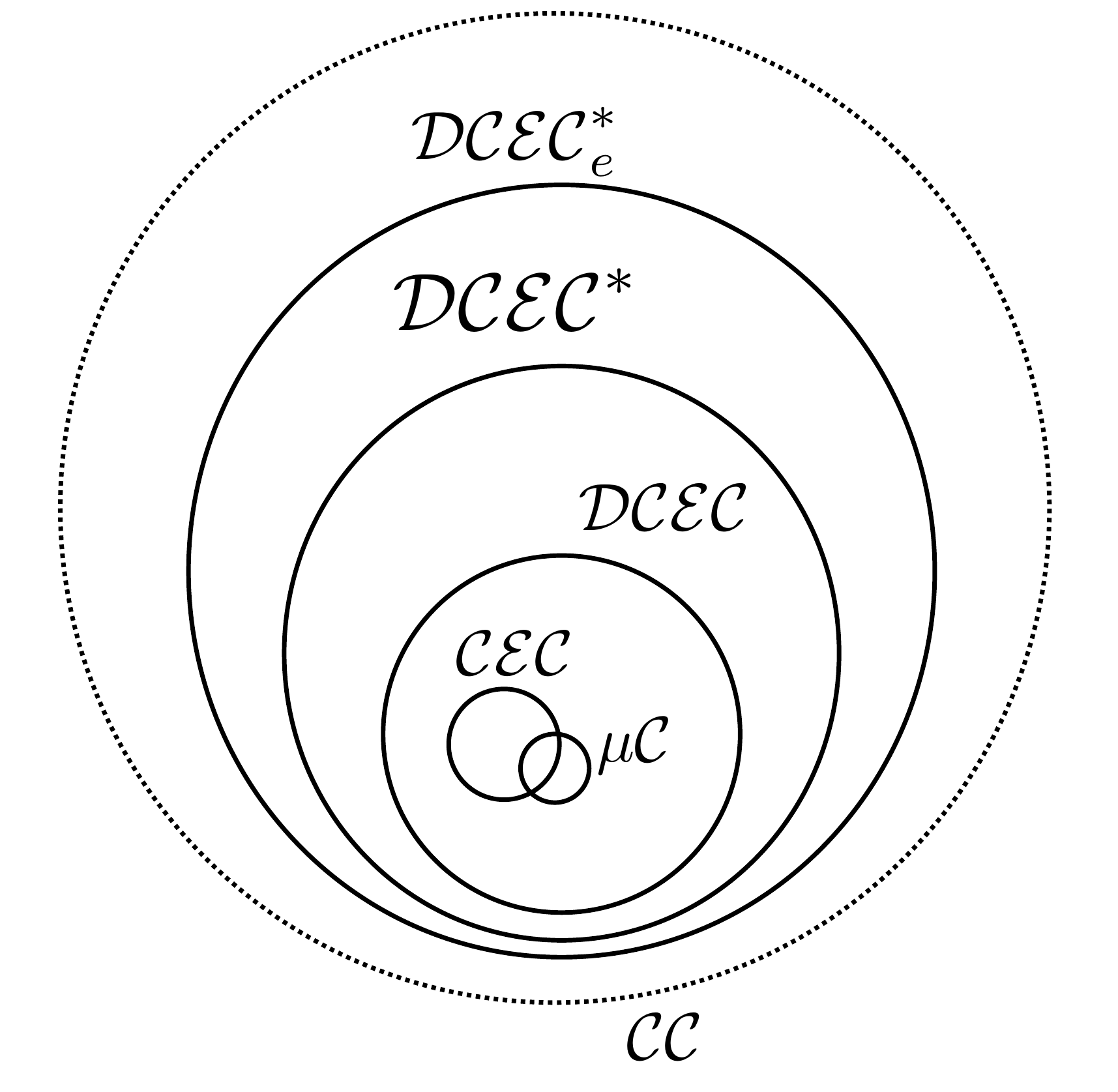}}
\caption{Cognitive Calculi. \textit{The \textbf{cognitive calculi}
    family is composed of a number of related
    calculi. \protect\citeauthor{ArkoudasAndBringsjord2008Pricai} introduced
    the first member in this family, $\mathcal{CEC}$, to model the
    false-belief task. The smallest member in this family,
    $\mu\mathcal{C}$, has been used to model uncertainty in quantified
    beliefs \protect\cite{govindarajulu2017strength}. \DCEC\ and variants have
    been used in the modelling of ethical principles and theories and
    their implementations.}}
 \label{fig:cc_family}
\end{figure}

\subsection{Syntax}
\label{subsect:syntax}

As mentioned above, \DCEC\ is a sorted calculus.  A sorted system can
be regarded as analogous to a typed single-inheritance programming
language.  We show below some of the important sorts used in \DCEC.\\
 
\begin{footnotesize}
\rowcolors{2}{gray!10}{white}
\def\arraystretch{1.25}

\begin{tabular}{lp{5.8cm}}  
\toprule
\textbf{Sort}    & \textbf{Description} \\
\midrule
\type{Agent} & Human and non-human actors.  \\

\type{Time} &  The \type{Time} type stands for
time in the domain.  E.g.\ simple, such as $t_i$, or complex, such as
$birthday(son(jack))$. \\

 \type{Event} & Used for events in the domain. \\
 \type{ActionType} & Action types are abstract actions.  They are
  instantiated at particular times by actors.  Example: eating.\\
 \type{Action} & A subtype of \type{Event} for events that occur
  as actions by agents. \\
 \type{Fluent} & Used for representing states of the world in the
  event calculus. \\
\bottomrule
\end{tabular}
\end{footnotesize} \\

The syntax has two components: a first-order
core and a modal system that builds upon this first-order core.  The
figures below show the syntax and inference schemata of \DCEC.    The first-order core of \DCEC\ is
the \emph{event calculus} \cite{mueller_commonsense_reasoning}.
Commonly used function and relation symbols of the event calculus are
included.  Fluents, event and times are the three major sorts of the event
calculus. Fluents represent states of the world as first-order
terms. Events are things that happen in the world at specific instants
of time. Actions are events that are carried out by an agent. For any
action type $\alpha$ and agent $a$, the event corresponding to $a$
carrying out $\alpha$ is given by $action(a, \alpha)$. For instance
if $\alpha$ is \textit{``running''} and $a$ is \textit{``Jack'' },
$action(a, \alpha)$ denotes \textit{``Jack is running''}.
Other calculi (e.g.\ the \emph{situation calculus}) for
modeling commonsense and physical reasoning can be easily switched out
in-place of the event calculus.

 \begin{scriptsize}
\begin{mdframed}[linecolor=white, frametitle=Syntax,
  frametitlebackgroundcolor=gray!10, backgroundcolor=gray!05,
  roundcorner=8pt]
 \begin{equation*}
 \begin{aligned} 
    \mathit{S} &::= 
    \begin{aligned}
      & \Agent \sep \ActionType \sep \Action \sqsubseteq
      \Event \sep \Moment  \sep \Fluent \\
    \end{aligned} 
    \\ 
    \mathit{f} &::= \left\{
    \begin{aligned}
      & action: \Agent \times \ActionType \rightarrow \Action \\
      &  \initially: \Fluent \rightarrow \Boolean\\
      &  \holds: \Fluent \times \Moment \rightarrow \Boolean \\
      & \happens: \Event \times \Moment \rightarrow \Boolean \\
      & \clipped: \Moment \times \Fluent \times \Moment \rightarrow \Boolean \\
      & \initiates: \Event \times \Fluent \times \Moment \rightarrow \Boolean\\
      & \terminates: \Event \times \Fluent \times \Moment \rightarrow \Boolean \\
      & \prior: \Moment \times \Moment \rightarrow \Boolean\\
    \end{aligned}\right.\\
        \mathit{t} &::=
    \begin{aligned}
      \mathit{x : S} \sep \mathit{c : S} \sep f(t_1,\ldots,t_n)
    \end{aligned}
    \\ 
    \mathit{\phi}&::= \left\{ 
    \begin{aligned}
     & q:\Boolean \sep  \neg \phi \sep \phi \land \psi \sep \phi \lor
     \psi \sep \forall x: \phi(x) \sep \\\
 &\perceives (a,t,\phi)  \sep \knows(a,t,\phi) \sep     \\ 
& \common(t,\phi) \sep
 \says(a,b,t,\phi) 
     \sep \says(a,t,\phi) \sep  \believes(a,t,\phi) \\
& \desires(a,t,\phi)  \sep \intends(a,t,\phi) \\ & \ought(a,t,\phi,(\lnot)\happens(action(a^\ast,\alpha),t'))
      \end{aligned}\right.
  \end{aligned}
\end{equation*}
\end{mdframed}
\end{scriptsize}

The modal operators present in the calculus include the standard
operators for knowledge $\knows$, belief $\believes$, desire
$\desires$, intention $\intends$, etc.  The general format of an
intensional operator is $\knows\left(a, t, \phi\right)$, which says
that agent $a$ knows at time $t$ the proposition $\phi$.  Here $\phi$
can in turn be any arbitrary formula. Also,
note the following modal operators: $\mathbf{P}$ for perceiving a
state, 
$\mathbf{C}$ for common knowledge, $\mathbf{S}$ for agent-to-agent
communication and public announcements, $\mathbf{B}$ for belief,
$\mathbf{D}$ for desire, $\mathbf{I}$ for intention, and finally and
crucially, a dyadic deontic operator $\mathbf{O}$ that states when an
action is obligatory or forbidden for agents. It should be noted that
\DCEC\ is one specimen in a \emph{family} of extensible
cognitive calculi.
 
The calculus also includes a dyadic (arity = 2) deontic operator
$\ought$. It is well known that the unary ought in standard deontic
logic leads to contradictions.  Our dyadic version of the operator
blocks the standard list of such contradictions, and
beyond.\footnote{A overview of this list is given lucidly in
  \cite{sep_deontic_logic}.}

Declarative communication of $\phi$ between $a$ and $b$ at time $t$ is
represented using the $\says(a,b,t, \phi)$.

\subsection{Inference Schemata}

The figure below shows a fragment of the inference schemata for \DCEC.
First-order natural deduction introduction and elimination rules are
not shown. Inference schemata $I_\mathbf{K}$ and $I_\mathbf{B}$ let us
model idealized systems that have their knowledge and beliefs closed
under the \DCEC\ proof theory.  While humans are not deductively
closed, these two rules lets us model more closely how more deliberate
agents such as organizations, nations and more strategic actors
reason. (Some dialects of cognitive calculi restrict the number of
iterations on intensional
operators.) 
$I_{13}$ ties intentions directly to perceptions
(This model does not take into account agents that could fail to carry
out their intentions).  $I_{14}$ dictates how obligations get
translated into known intentions.

\begin{scriptsize}

\begin{mdframed}[linecolor=white, frametitle=Inference Schemata
  (Fragment), nobreak=true, frametitlebackgroundcolor=gray!10, backgroundcolor=gray!05, roundcorner=8pt]
\begin{equation*}
\begin{aligned}
  &\hspace{40pt} \infer[{[I_{\knows}]}]{\knows(a,t_2,\phi)}{\knows(a,t_1,\Gamma), \ 
    \ \Gamma\vdash\phi, \ \ t_1 \leq t_2}  \\ 
& \hspace{40pt} \infer[{[I_{\believes}]}]{\believes(a,t_2,\phi)}{\believes(a,t_1,\Gamma), \ 
    \ \Gamma\vdash\phi, \ \ t_1 \leq t_2} \\
& \hspace{20pt} \infer[{[I_4]}]{\phi}{\knows(a,t,\phi)}
\hspace{18pt}\infer[{[I_{13}]}]{\perceives(a,t', \psi)}{t<t', \ \ \intends(a,t,\psi)}\\
&\infer[{[I_{14}]}]{\knows(a,t,\intends(a,t,\chi))}{\begin{aligned}\ \ \ \ \believes(a,t,\phi)
 & \ \ \
 \believes(a,t,\ought(a,t,\phi, \chi)) \ \ \ \ought(a,t,\phi,
 \chi)\end{aligned}}
\end{aligned}
\end{equation*}
\end{mdframed}
\end{scriptsize}

 \subsection{Semantics}

 The semantics for the first-order fragment is the standard
 first-order semantics. The truth-functional connectives
 $\land, \lor, \rightarrow, \lnot$ and quantifiers $\forall, \exists$
 for pure first-order formulae all have the standard first-order
 semantics. The semantics of the modal operators differs from what is
 available in the so-called Belief-Desire-Intention (BDI) logics
 {\cite{bdi_krr_1999}} in many important ways.  For example, \DCEC\
 explicitly rejects possible-worlds semantics and model-based
 reasoning, instead opting for a \textit{proof-theoretic} semantics
 and the associated type of reasoning commonly referred to as
 \textit{natural deduction}
 \cite{gentzen_investigations_into_logical_deduction,proof-theoretic_semantics_for_nat_lang}.
 Briefly, in this approach, meanings of modal operators are defined
 via arbitrary computations over proofs.


\section{Defining TAI}
\begin{figure}[h!]
 \centering
 {
  \includegraphics[width=0.75\linewidth]{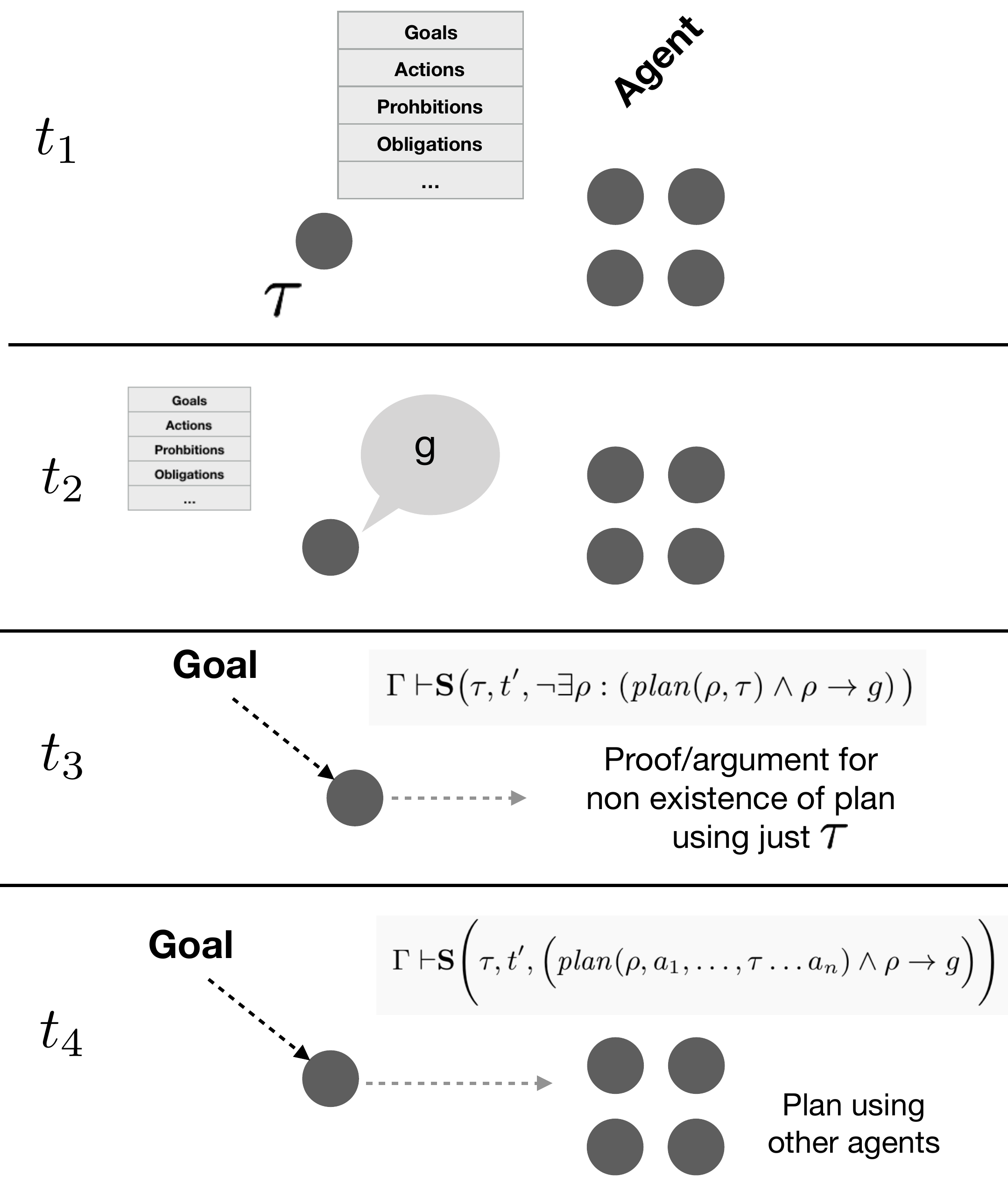}}
\caption{TAI Working Through Time. \textit{A TAI agent initially
    considers a goal and then has to produce a proof for the
    non-existence of a non-tentacular plan that uses only this
    agent. Then $\tau$ recruits a set of other relevant agents to help
    with its goal.}}
 \label{fig: TAI_Architecture}
\end{figure}

We denote the state-of-affairs at any time $t$ by a set of formulae
$\Gamma(t)$. This set of formulae will also contain any obligations
and prohibitions on different agents. For each agent $a_i$ at time
$t$, there is a contract $\mathbf{c}(a_i, t)\subseteq \Gamma(t)$ that
describes $a_i$'s obligations, prohibitions etc.  $a$ at any time $t$
then comes up with a goal $g$ so that its
contract is satisfied.\footnote{See \cite{nsg_sb_dde_2017} for an
  example of how obligations and prohibitions can be used in \DCEC.}
The agent believes that if  $g$ does not hold then its contract at some future
$t+\delta$  will be violated:

 $$\mathbf{B}\left(a, t, \lnot
 g \rightarrow  \lnot \bigwedge \mathbf{c}(a, t+\delta)\right)$$ Then the agent tries
 to come up with a plan involving a sequence of actions to satisfy the
 goal.

We make these notions more precise. An agent $a$ has a set of
actions that it \emph{can} perform at different time points. For instance, a
vacuuming agent can have movement along a plane as its possible
actions while an agent on a phone can have displaying a notification
as an action. We denote this by $can(a,\alpha, t)$ with the following
additional axiom:

$$ \colorbox{gray!10}{Axiom}  \lnot can(a,\alpha,
t) \rightarrow \lnot \happens(action(a, \alpha), t) $$

\noindent We now define a \emph{consistent plan} below:

 \begin{footnotesize}
\begin{mdframed}[linecolor=white, frametitle=Consistent Plan,
  frametitlebackgroundcolor=gray!10, backgroundcolor=gray!05,
  roundcorner=8pt]

  A \emph{consistent plan} $\rho_{\langle a_1, \ldots, a_n\rangle} $ at
  time $t$ is a sequence of agents $a_1,\ldots, a_n$ with
  corresponding actions $\alpha_1, \ldots, \alpha_n$ and times
  $t_1, \ldots, t_n$ such that
  $\Gamma \vdash (t<t_i < t_j) \mbox{ for } i<j$ and for all agents $a_i$
  we have:
\begin{enumerate} 
\item $can(a_i, \alpha_i, t_i)$
\item $happens(action(a_i,\alpha_i))$ is consistent with $\Gamma(t)$. 
\end{enumerate}
 \end{mdframed}
 \end{footnotesize}
 Note that a consistent plan $\rho_{\langle \ldots \rangle}$ can be
 represented by a term in our language. We introduce a new sort
 $\mathsf{Plan}$ and a variable-arity predicate symbol
 $\plan(\rho, a_1, \ldots, a_n)$ which says that $\rho$ is a plan
 involving $a_1\ldots, a_n$.

 A goal is also any formula $g$.  A consistent plan satisfies a goal
 $g$ if:

\begin{equation*}
 \left(\begin{aligned}
\Gamma(t) \cup \left\{ \begin{aligned}
   & happens(action(a_1,\alpha_1), t_1), \ldots,\\
 &happens(action(a_n,\alpha_n), t_n) \end{aligned}\right\} 
 \end{aligned}\right) \vdash g
 \end{equation*}
 We use $\Gamma\vdash (\rho \rightarrow g)$ as a shorthand for the
 above. The above definitions of plans and goals give us the following
 important constraint needed for defining TAI. This differentiates our
 planning formalism from other planning systems and makes it more
 appropriate for an architecture for a general-purpose tentacular AI
 system.

 \begin{footnotesize}
\begin{mdframed}[linecolor=white, frametitle=Uniform Planning Constraint,
  frametitlebackgroundcolor=yellow!25, backgroundcolor=yellow!10,
  roundcorner=8pt]
  Plans and goals should be represented and reasoned over in the
  language of the planning system.
\end{mdframed}
 \end{footnotesize}

 Leveraging the above requirement, we can define two levels of TAI
 agents. A $\level{1}{*}$ TAI system corresponding to an agent $\tau$
 is a system that comes up with goal $g$ at time $t'$ to satisfy its
 contract, produces a proof that there is no consistent plan that
 involves only the agent $\tau$. Then $\tau$ comes with a plan that
 involves one or more other agents.  A $\level{1}{*}$ TAI agent starts
 with knowledge about what actions are possible for other agents.

 \begin{footnotesize}
   \begin{mdframed}[linecolor=white, frametitle=$\level{1}{*}$ TAI Agents ,
     frametitlebackgroundcolor=gray!10, backgroundcolor=gray!05,
     roundcorner=8pt]
     \begin{enumerate}
       \item[\textbf{Prerequisite}] For any $a$, $\alpha$, $t$, we have:
         \begin{equation*}
           \begin{aligned}
             \Gamma \vdash
             & can(a, \alpha, t) \rightarrow \Knows\big(\tau, t',
             can(a, \alpha, t) \big) \end{aligned}
         \end{equation*}
\item[\textbf{Then}]
       \item $\tau$ produces a proof that no plan exists for $g$
         involving just itself and $\tau$ declares that there is no
         such plan.
         \begin{equation*}
           \begin{aligned}
             \Gamma \vdash
             & \says\big(\tau, t', \lnot \exists  \rho:\left( \plan(\rho,
             \tau) \land \rho \rightarrow g\right)\big) \end{aligned}
         \end{equation*}
       \item $\tau$ produces a plan for $g$ involving just itself and
         one or more agents and declares that plan.
         \begin{equation*}
           \begin{aligned}
             \Gamma \vdash
             & \says\Bigg(\tau, t', \Big( \plan(\rho
            , a_1, \ldots, \tau \ldots a_n) \land \rho \rightarrow g\Big)\Bigg) 
         \end{aligned}
         \end{equation*}
       \end{enumerate}
     \end{mdframed}
   \end{footnotesize}

   The agent may not always have knowledge about what other agents can
   do. The TAI agent may have imperfect knowledge about other
   agents. The agent can gain information about other agents' actions,
   their obligations, prohibitions, etc. by observing them or by reading
   specifications governing these agents. In this case, we get a
   $\level{2}{*}$ TAI agent. We need to modify only the prerequisite
   condition above.

 \begin{footnotesize}
   \begin{mdframed}[linecolor=white, frametitle=$\level{2}{*}$ TAI Agents ,
     frametitlebackgroundcolor=gray!10, backgroundcolor=gray!05,
     roundcorner=8pt]
     \begin{enumerate}
     \item[\textbf{Prerequisite}]  For any $a$, $\alpha$, $t$, we have:
         \begin{equation*}
           \begin{aligned}
             \Gamma \vdash
             & can(a, \alpha, t) \rightarrow \Believes\big(\tau, t',
             can(a, \alpha, t) \big) \end{aligned}
         \end{equation*}
      
       \end{enumerate}
     \end{mdframed}
   \end{footnotesize}
 
   The TAI agents above can be considered \textbf{first-order}
   tentacular agents. We can also have a \textbf{higher-order} TAI
   agent that intentionally engages in actions that trigger one or
   more other agents to act in tentacular fashion as described above.
   The need for having the uniform planning constraint is more clear
   when we consider higher-order agents.


\section{A Hierarchy of TAI Agents}
The TAI formalization above gives rise to multiple hierarchies of
tentacular agents.  We discuss some of the these below.

\begin{small}
\begin{enumerate}
\item[\textbf{Syntactic Goal Complexity}] The goal $g$ can range in
  complexity from simple propositional statements,
  e.g. $clean{Kitchen}$, to first-order statements.  e.g.
  $\forall r:\mathsf{Room}: clean(r)$, and to intensional statements
  representing cognitive states of other agents
  $$\believes(a, now, \believes(b, now,\forall r: clean(r)))$$


\item[\textbf{Goal Variation}] According to the definition
  above, an agent $a$ qualifies as being tentacular if it plans for
  just one goal $g$ in tentacular fashion as laid out in the
  conditions above. We could have agents that plan for a number of
  varied and different goals in tentacular fashion. 


\item[\textbf{Plan Complexity}] For many goals, there will usually be
  multiple plans involving different actions (with different costs and
  resources used) and executed by different agents. 


\end{enumerate}
\end{small}

\begin{figure}[h!]
 \centering
 {
  \includegraphics[width=\linewidth]{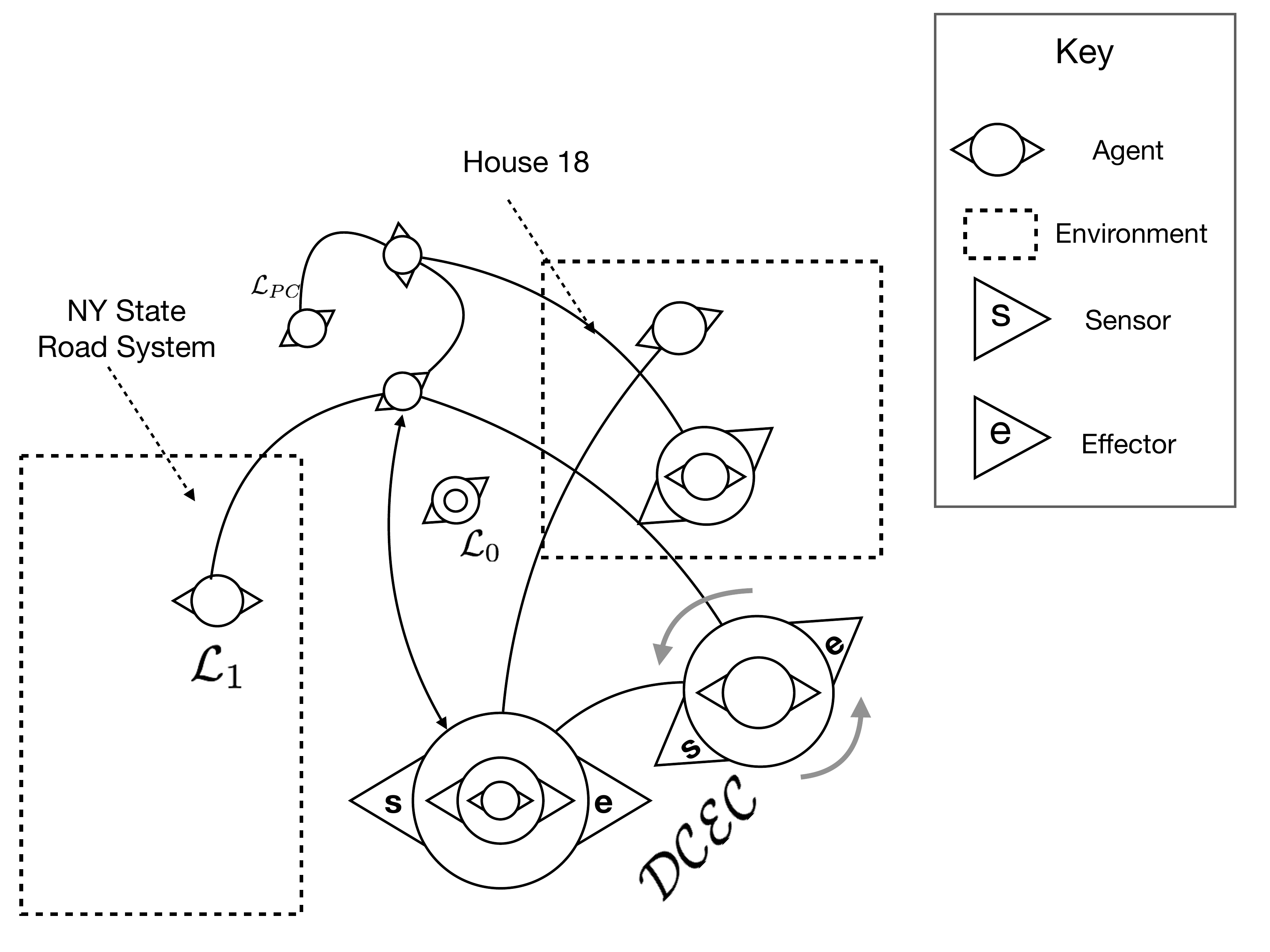}}
 \caption{Pictorial Overview.  \textit{A bit of explanation: That some
     agents are within agents indicates that the outer agent knows
     and/or believes everything relevant about the inner agent; hence
     as agents are increasingly cognitively powerful, the depth of
     their epistemic attitudes grows (reflected in formulae with
     iterated belief/knowledge operators).  Agents grow in
     size/intelligence in lockstep with the logical calculi upon which
     they are based increasing in expressivity and reasoning power;
     $\mathcal{L}_0$ is zero-order logic, $\mathcal{L}_1$ is
     e.g.\ first-order logic, and the particular cognitive calculus
     $\mathcal{DCEC}$ is shown.  Rotation indicates simply that,
     through time, agents perceive and act.}}
 \label{fig:pictorial_overview}
\end{figure}


\section{Examples and Embryonic Implementation}
In this section, we present a formal sketch of a TAI agent and then
describe using another example ongoing work in implementing a TAI
system.

\subsection{Example } Consider the example given in the beginning.  We
have a human $j$ and three artificial agents: $a_{c}$ in the car,
$a_h$ in the home and $a_p$ an agent managing scheduling and calendar
information.  We present some of the formulae in $\Gamma$.

\begin{footnotesize}
\begin{equation*}
\begin{aligned}
& \mathbf{B}(a_c, t_0, crowded(store) \rightarrow unusal), \mbox{\colorbox{gray!20}{$\mathsf{f_1}$}}\\
&\mathbf{P}(a_c, t_1, crowded(store)),\mbox{\colorbox{gray!20}{$\mathsf{f_2}$}}\\
&\forall t: \mathbf{O}\left(\begin{aligned}a_c, t, & unusal,\\ &
    happens\big(action(a_c, check(weather)),
    t+1\big) \end{aligned}\right)\mbox{\colorbox{gray!20}{$\mathsf{f_3}$}}\\
&  \forall t: \mathbf{B}\left(a_c, t, \mbox{\colorbox{gray!20}{$\mathsf{f_3}$}}\right)\\
& \forall a: \left(\begin{aligned} happens\Big(& action(a, check(weather)), t_3\Big) 
\\ & \rightarrow\mathbf{K}(a, t_4, storm), \end{aligned}\right)\mbox{\colorbox{gray!20}{$\mathsf{f_4}$}}\\
&\forall t: \mathbf{O}\big(a_c, t, storm, \mathsf{S}(a_c, a_h, storm,
t+1))\big),\mbox{\colorbox{gray!20}{$\mathsf{f_5}$}}\\
&  \forall t: \mathbf{B}\left(a_c, t, \mbox{\colorbox{gray!20}{$\mathsf{f_5}$}}\right)\\
 \end{aligned} 
\end{equation*}
\end{footnotesize}
The above formulae first state the fact that $a_c$ observes the store
being crowded. $a_c$'s contract states that the agent should check a
weather service if it finds something unusual.  The formulae also
states that if an agent checks the weather at $t_3$, the agent will
get a prediction about an incoming storm.  $a_c$'s contract places an
obligation on it to inform $a_h$ if it believes that a storm is
incoming.
 
\begin{footnotesize}
\begin{equation*}
\begin{aligned}
&\forall t: \mathbf{O}\big(a_h, t, storm, \forall s: quantity(s) > 0\big),\mbox{\colorbox{gray!20}{$\mathsf{f_6}$}}\\
& \mathbf{K}\left(\begin{aligned} a_h, t_5, & shops(j, today) \lor shops(j,
tomorrow) \\ &  \rightarrow \forall s: quantity(s) > 0\end{aligned} \right), \mbox{\colorbox{gray!20}{$\mathsf{f_7}$}}\\
& \forall t: \mathbf{B}\left(\begin{aligned}a_h, t, & happens\big(action(a_c, recc(shops(j))), t\big)\\ & \rightarrow shops(j)\end{aligned}\right) \mbox{\colorbox{gray!20}{$\mathsf{f_8}$}}\\
& \forall t: \mathbf{B}\left(\begin{aligned} & a_h, t, happens\Big(action\big(a_h, req(a_c, shops(j))\big), t\Big)\\ & \rightarrow happens\big(action(a_c, recc(shops(j))), t\big)\end{aligned}\right) \mbox{\colorbox{gray!20}{$\mathsf{f_9}$}}\\
 \end{aligned} 
\end{equation*}
\end{footnotesize}

The first formula above states that $a_h$ ought to see to it that
supplies are stocked in the event of a storm. Then we have that 
$a_h$ knows that the human $j$ shopping today or tomorrow can result
in the supplies being stocked. $a_h$ gets information from $a_p$ that
shopping tomorrow is not possible (this formula is not shown). Then we
have formulae stating the effects of $a_c$ recommending the shopping
action to $j$.  The goal for $a_h$ is $\forall s: quantity(s)>0$ and a
plan for it is built up using $a_h$, $a_c$ and $j$.


\subsection{Toward an Implementation}
\label{subsect:implementation}

We describe an example scenario that we are targeting for an
embryonic implementation.

Beforehand, a number of contracts have been executed that bind the
adult parents $P_1$ and $P_2$ in a home $H$, and also bind a number of
artificial agents in $H$, including a TAI agent ($\tau$) that oversees
the home.  (Strictly speaking, the agents wouldn't have entered into
contracts, but they would know that their human owners have done so,
and they would know what the contracts are.)

\begin{quote}{It’s winter in Berlin NY.  Night.  Outside, a blizzard.  The
  mother and father of the home $H$, and their two toddler children,
  are fast asleep.  The smartphone of each parent is set to ``Do Not
  Disturb'', with incoming clearance for only close family.  There is
  no landline phone.  A carbon monoxide sensor in the basement, near
  the furnace, suddenly shows a readout indicating an elevated level,
  which proceeds to creep up.  $\tau$ perceives this, and forms
  hypotheses about what is causing the elevated reading, and believes
  on the basis of using a cognitive calculus that the reading is
  accurate (to some likelihood factor).  The nearest firehouse is
  notified by $\tau$ .  No alarm sounds in the house.  $\tau$ runs a
  diagnostic and determines that the battery for the central auditory
  alarm is shot.  The reading creeps up higher, and now even the
  sensors in the upstairs bedrooms where the humans are asleep show an
  elevated, and climbing, level.  $\tau$ perceives this too.}
\end{quote}

Unfortunately, $\tau$ reasons that by the time the firemen arrive,
permanent neurological damage or even death may well (need again a
likelihood factor) be caused in the case of one or more members of the
family.  Should the alarm company have programmed the sensor to report
to a central command, still, any human command is fallible. The
company may be negligent, or a phone call may be the only option at
their disposal, or they may dispatch personnel who arrive too
late. Without enlisting the help of other \textit{artificial} agents
in planning and reasoning, $\tau$ can't save the family; $\tau$ knows
this on the basis of proof/argument.

However, $\tau$ can likely wake the family up, starting with the
parents, in any number of ways.  However, each of these ways entails
violation of at least one legal prohibition that has been created by
contracts that are in place.  These contracts have been analyzed by an
IBM service, which has stocked the mind of $\tau$ with knowledge of
legal obligations in \DCEC --- or rather in a dialect that has
separate obligation operators for legal $\ought_l$ and moral
$\ought_m$ obligations.  The moral obligation to save the family
overrides the legal prohibitions, however.  $\tau$ turns on the TV in
the master bedroom at maximum volume, and flashes a warning to leave
the house immediately because of the lethal gas building up.  (There
are many other alternatives, of course.  TAI could break through Do
Not Disturb, eg).

\subsection{Toward Using Smart-City Infrastructure}
\label{subsect:toward_smartcity}

The European Initiative on Smart Cities \cite{europa} is an effort by
the European Commission \cite{ec} to improve the quality of life
throughout Europe, while progressing toward energy and climate
objectives.  Many of its goals are relevant to and desirable in the
world at large.  TAI has the potential to be instrumental in achieving
many of these, such as smart appliances (in the manner discussed in
the previous sub-section) and intelligent traffic management.  Indeed,
the scope and objectives of the Initiative may conceivably be
considerably broadened with a pervasive application of TAI.

We briefly point at a simple scenario that expands on the vision of
the European Initiative's smart-transportation goals.

\begin{quote}{Parking space is very scarce on a work-day in mid-town
  Manhattan.  A busy executive will need to park near several offices
  over the course of the day, and these locations change over the
  week.}
\end{quote}

The executive's car consults her calendar.  Based on past patterns, it
interpolates locations where it believes she intends to park.  It
communicates with other cars parked at these locations, and determines
when their owners are likely to return, based on their expressed (and
inferable) intentions and current locations.  Adjusting for the
location of our executive, traffic conditions and changes in her
agenda, it determines the optimal parking locations dynamically,
throughout her busy day.  Of course, in the spirit of TAI, all other
cars would have their movement adjusted accordingly, through
time.\footnote{TAI applications like this give rise to privacy concerns which could possibly
  be resolved by employing either \textbf{differential privacy} 
  \cite{dwork2008differential} or privacy based on
  \textbf{zero-knowledge proofs} \cite{gehrke2011towards}.}


\section{Conclusion \& Future Work}
\label{sect:conclusion}
We have introduced Tentacular AI, and a number of architectural
elements thereof, and are under no illusion that we have accomplished
more than this.  At AEGAP 2018, we will demonstrate TAI in action in
both the scenarios sketched above; implementation is currently
underway.  Despite the nascent state of the TAI research program, we
hope to have provided a promising, if inchoate, overview of tentacular
AI --- an overview which, given the centrality of highly expressive
languages for novel planning and reasoning, we hope is of interest to
some, maybe even many, at this dawn of the ``internet of things'' and
its vibrant intersection with AI.


\section{Acknowledgments}
\label{sect:ack}

The TAI project is made possible by joint support from RPI and IBM
under the AIRC Program; we are grateful for this support.  Some of the
research reported on herein has been enabled by support from ONR and
AFOSR, and for this too we are grateful.

 \bibliographystyle{named}
\bibliography{main72,naveen,kartik,atriya}

\end{document}